\documentclass[10pt,twocolumn,letterpaper]{article}

\usepackage[pagenumbers]{wacv} 

\usepackage{graphicx}
\usepackage{amsmath}
\usepackage{amssymb}
\usepackage{booktabs}
\usepackage[accsupp]{axessibility}

\usepackage[pagebackref,breaklinks,colorlinks]{hyperref}

\usepackage[capitalize]{cleveref}
\crefname{section}{Sec.}{Secs.}
\Crefname{section}{Section}{Sections}
\Crefname{table}{Table}{Tables}
\crefname{table}{Tab.}{Tabs.}

\def\wacvPaperID{1878} 
\def\confName{WACV}
\def\confYear{2025}

\usepackage{multirow}
\usepackage{array}
\usepackage{booktabs}
\usepackage[table]{xcolor}
\definecolor{tabfirst}{rgb}{0.4, 0.65, 0.3} 
\definecolor{tabsecond}{rgb}{0.7, 0.8, 0.65} 

\begin{document}

\title{Cascaded Dual Vision Transformer for Accurate Facial Landmark Detection}

\author{Ziqiang Dang$^{1,2,*}$, Jianfang Li$^{2,*,\dag}$, 
Lin Liu$^{2}$\\
$_{}^{1}\textrm{}$Zhejiang University, $_{}^{2}\textrm{}$Institute for Intelligent Computing, Alibaba Group\\
}


\maketitle

\renewcommand{\thefootnote}{\fnsymbol{footnote}}
\footnotetext[1]{Ziqiang Dang and Jianfang Li contributed equally to this work.}
\footnotetext[2]{Corresponding author: Jianfang Li, wuhui.ljf@alibaba-inc.com.}

\begin{abstract}
Facial landmark detection is a fundamental problem in computer vision for many downstream applications. This paper introduces a new facial landmark detector based on vision transformers, which consists of two unique designs: Dual Vision Transformer (D-ViT) and Long Skip Connections (LSC). Based on the observation that the channel dimension of feature maps essentially represents the linear bases of the heatmap space, we propose learning the interconnections between these linear bases to model the inherent geometric relations among landmarks via channel-split ViT. We integrate such channel-split ViT into the standard vision transformer (\textit{i.e.}, spatial-split ViT), forming our Dual Vision Transformer to constitute the prediction blocks. We also suggest using long skip connections to deliver low-level image features to all prediction blocks, thereby preventing useful information from being discarded by intermediate supervision. Extensive experiments are conducted to evaluate the performance of our proposal on the widely used benchmarks, \textit{i.e.}, WFLW~\cite{Wu_2018_CVPR}, COFW~\cite{Burgos-Artizzu_2013_ICCV}, and 300W~\cite{Sagonas_2013_ICCV_Workshops}, demonstrating that our model outperforms the previous \mbox{SOTAs} across all three benchmarks.
\end{abstract}

\section{Introduction}

Facial landmark detection involves locating a set of predefined key points on face images, serving as a fundamental step for supporting various high-level applications, including face alignment~\cite{2015Efficient,Kazemi2014One,2015Face,2015Face}, face recognition~\cite{2013Random,Zhu_2015_CVPRface,WANG2021215}, face parsing~\cite{Smith_2013_CVPR} and 3D face reconstruction~\cite{Deng_2019_CVPR_Workshops,Danecek_2022_CVPR,li2023robust}.

Early approaches of landmark detection relied on statistical models~\cite{927467,670965}, but have been surpassed by modern landmark detectors that use convolutional neural networks (\ie, CNN)~\cite{Sun_2013_CVPR, Zhou_2013_ICCV_Workshops,Zhu_2015_CVPR}.  CNNs learn a transformation between image features and a set of 2D coordinates or regress heatmaps to represent the probability distribution of landmarks. Besides, these detectors~\cite{Dapogny_2019_ICCV,Wang_2019_ICCV,huang2021adnet, DBLP:journals/corr/abs-2104-03100, Zhou_2023_CVPR} usually employ cascaded networks in conjunction with intermediate supervision to progressively refine the predictions, producing remarkable advances. 
To capture the intrinsic geometric relationships among landmarks for accurate predictions, researchers have also developed various effective methods using Graph Convolutional Networks (GCNs)~\cite{9442331, Li2020StructuredLD}, Graph Attention Networks (GATs)~\cite{Prados-Torreblanca_2022_BMVC} or Transformers~\cite{yang2021transpose, 9878969, Xia_2022_CVPR}. 
However, these methods primarily utilize patch-based image features to learn spatial relations among landmarks. 

In this paper, we have developed a vision transformer-based model architecture and modeled the intrinsic geometric relationships among landmarks by computing the correlations between the linear bases of heatmap space, enabling us to achieve new SOTA results across all three benchmarks.
Following the design paradigm of cascaded networks~\cite{Dapogny_2019_ICCV,Wang_2019_ICCV,huang2021adnet, DBLP:journals/corr/abs-2104-03100, Zhou_2023_CVPR}, we repeat prediction blocks in conjunction with intermediate supervisions. Specially, in our architecture, we propose two unique designs: Dual Vision Transformer (D-ViT), which constitutes the prediction blocks, and Long Skip Connections (LSC), the strategy for connecting these blocks.

The standard vision transformer~\cite{50650} discretizes images or feature maps into small patches, then rearranges them into a sequence to extract global and local image features. However, it lacks the ability to model the underlying geometric characteristics of landmarks. To address this, we propose incorporating the channel-split ViT to model the inherent relationships among landmarks. Specifically, the prediction block outputs a feature map $F \in \mathbb{R}^{C\times H \times W}$ for intermediate supervision, which can be split along the channels into $F=(f_1,f_2,...,f_C)$. Therefore, when considering using convolution operations to regress the feature map $F$ into the heatmaps, the heatmap for each landmark is actually a linear combination of $f_m$. To put it another way, the channel dimension of feature maps essentially represents the linear bases of the heatmap space. Based on such insight, we take advantage of the transformer architecture to learn underlying relationships among these linear bases, allowing for adaptive computation of their interconnections through the multi-head self-attention mechanism. Finally, the spatial-split ViT and the channel-split ViT together form our Dual Vision Transformer (D-ViT).


Following the classic stacked Hourglasses networks~\cite{Newell2016StackedHN,Wang_2019_ICCV,huang2021adnet,Zhou_2023_CVPR}, which utilize residual connections between two sequential hourglass architectures, we first also apply the same connection strategy to boost the network. However, we find that when the number of prediction blocks exceeds 4, the detection performance is instead diminished by deeper model architecture. As far as we know, this is caused by intermediate supervisions, which can lead to the inadvertent discard of useful information. To handle this problem, we suggest using long skip connections to deliver low-level image features to all prediction blocks, thereby making deeper network architectures feasible.

We evaluate the performance of our proposal on the widely used benchmarks, \ie, WFLW~\cite{Wu_2018_CVPR}, COFW~\cite{Burgos-Artizzu_2013_ICCV}, and 300W~\cite{Sagonas_2013_ICCV_Workshops}. Extensive experiments
demonstrate that our approach outperforms the previous state-of-the-art methods and achieves a new SOTA across all three datasets. The main contributions can be summarized as follows:

1) We introduce a facial landmark detector based on our unique dual vision transformer (D-ViT), which is able to effectively capture contextual image features and underlying geometric relations among landmarks via spatial-split and channel-split features. 

2) To avoid losing useful information due to intermediate supervision and make deeper network architectures feasible, we propose a unique connection strategy, \ie, Long Skip Connections, to transmit low-level image features from ResNet to each prediction block.

3) Extensive experiments are conducted to evaluate our approach, demonstrating its good generalization ability and superior performance compared to existing SOTAs across three publicly available datasets (\ie, WFLW, COFW, and 300W). Our code will be released for reproduction.

\section{Related Work}
In the literature on facial landmark detection, deep learning methods can generally be divided into two categories: coordinate regression-based method and heatmap-based method.

\textbf{Coordinate regression-based} methods directly regress facial landmarks through learning the transformation between image features and landmark coordinates. These methods~\cite{Sun_2013_CVPR, Zhou_2013_ICCV_Workshops,Zhu_2015_CVPR, Trigeorgis_2016_CVPR, Lv_2017_CVPR, Dapogny_2019_ICCV} are typically designed in a coarse-to-fine manner, employing multiple prediction stages or cascaded network modules to gradually refine the landmark coordinates. 
In these methods, ResNet~\cite{He_2016_CVPR} combined with wing loss~\cite{Feng_2018_CVPR} is commonly used as the backbone to extract image features and regress landmark coordinates.
DTLD~\cite{9878969} adopts pretrained ResNet~\cite{He_2016_CVPR}  to extract multi-scale image features and apply cascaded transformers to gradually refine landmarks by predicting offsets.
Considering the underlying geometric relationships among landmarks, SDL~\cite{Li2020StructuredLD} and SDFL~\cite{9442331} utilize graph convolutional networks to explicitly capture structural features.
Besides, SLPT~\cite{Xia_2022_CVPR} introduces a sparse local patch transformer to learn the intrinsic landmark relations, which extracts the representation (\ie, embedding) of each individual landmark from the corresponding local image patch and processes them based on the attention mechanism. 

\textbf{Heatmap-based} methods~\cite{dong2018san, 8641467, Newell2016StackedHN} predict a heatmap for each landmark, where the point with the highest intensity, or its vicinity, is considered the optimal position of the landmark.
In these methods, UNet~\cite{RFB15a} and the stacked hourglasses network~\cite{Newell2016StackedHN} are frequently used as backbone architectures.
HRNet~\cite{sun2019deep, WangSCJDZLMTWLX19} showed remarkable results through the combination of multi-scale image features. Adaptive wing loss~\cite{Wang_2019_ICCV} was proposed as the loss function for heatmap regression to balance the normal and hard cases. Besides, the predictions can be further improved by integrating coordinate encoding with CoordConv~\cite{Liu2018AnIF}. However, heatmap-based methods usually suffer from discretization-induced errors, since the heatmap size is usually much smaller than the input image. Consequently, various methods have been developed to alleviate discretization-induced errors, including the usage of heatmap matching to improve accuracy~\cite{NIPS2014_e744f91c}, continuous heatmap encoding and decoding method~\cite{bulat2021subpixel}, differential spatial to numerical transform (DSNT)~\cite{DBLP:journals/corr/abs-1801-07372} and Heatmap in Heatmap (HIH)~\cite{DBLP:journals/corr/abs-2104-03100} for subpixel coordinates.
Additionally, LAB~\cite{Wu_2018_CVPR} suggests predicting the facial boundary as a geometric constraint to help regress the landmark coordinates. LUVL~\cite{kumar2020luvli} predicts not only the landmark positions but also the uncertainty and probability of visibility for better performance. SPIGA~\cite{Prados-Torreblanca_2022_BMVC} combines CNN with cascaded Graph Attention Networks to jointly predict head pose and facial landmarks.  ADNet~\cite{huang2021adnet} introduces anisotropic direction loss (ADL) and anisotropic attention module (AAM) to address ambiguous landmark labeling. STARLoss~\cite{Zhou_2023_CVPR} adaptively suppresses the prediction error in the first principal component direction to mitigate the impact of ambiguity annotation during the training phase. LDEQ~\cite{Micaelli_2023_CVPR} employs Deep Equilibrium Models~\cite{bai2019deep} to detect face landmarks and achieves state-of-the-art results on the WFLW benchmark~\cite{Wu_2018_CVPR}. Recently, FRA~\cite{gao2024self} learned a general self-supervised facial representation for various facial analysis tasks, achieving state-of-the-art results on the 300W benchmark~\cite{Sagonas_2013_ICCV_Workshops} among self-supervised learning methods for facial landmark detection. Most of these methods utilize convolutional neural networks and produce remarkable results. 
In this work, we have developed a vision transformer-based model architecture and achieved new SOTA results across all three benchmarks.



\begin{figure*}[htbp]
    \centering
    \includegraphics[width=0.95\linewidth]{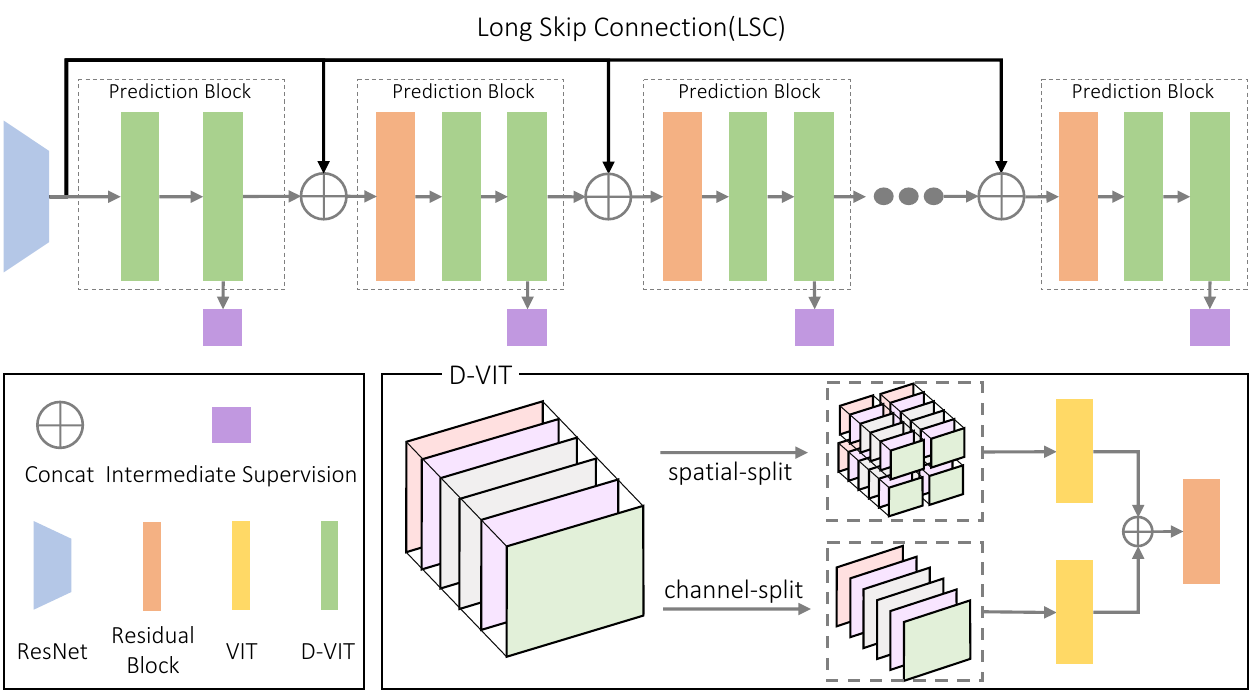}
    \vspace{-0.2cm}
    \caption{Architecture of our framework. 
    Long skip connections (depicted as upper black lines) distribute low-level features to each prediction block, thereby preventing useful information from being discarded by intermediate supervision.
    D-ViT learns contextual features of the image through spatially split patches and captures the underlying geometric relations among landmarks using channel-split features.
    }
    \label{fig:model}
\end{figure*}

\section{Method}
The architecture of our proposed model is presented in~\cref{fig:model}, which utilizes ResNet~\cite{He_2016_CVPR} to extract low-level image features and employs repeated prediction blocks in conjunction with intermediate supervision to gradually improve the detection performance. 
We will describe the core design of our architecture, Dual Vision Transformer and Long Skip Connection, in~\cref{Subsec:DVIT} and~\cref{Subsec:LSP}, respectively, and then introduce the training loss in~\cref{Subsec:Loss}.

\subsection{Dual Vision Transformer}
\label{Subsec:DVIT}
Our dual vision transformer (D-ViT) is built upon the standard vision transformer (ViT)~\cite{50650}, which discretizes the input image or feature map into smaller spatial patches and arranges these patches into a sequence. Then, the attention mechanism is utilized to establish relationships among the patches, allowing for the extraction of both local and global image features. However, the standard ViT does not explicitly leverage the geometric relationships among landmarks. 
Consequently, we propose to incorporate a channel-split ViT to establish the relationships between bases in the heatmap space, thereby extracting the underlying geometric features among landmarks.
The proposed channel-split ViT can be seamlessly integrated into the ViT architecture without the need for extra steps, such as explicit conversion to the heatmaps or landmark coordinates. Finally, the spatial-split ViT and the channel-split ViT together form our Dual Vision Transformer (D-ViT), as shown in~\cref{fig:model}.

\begin{figure}[!ht]
    \centering
    \includegraphics[width=\linewidth]{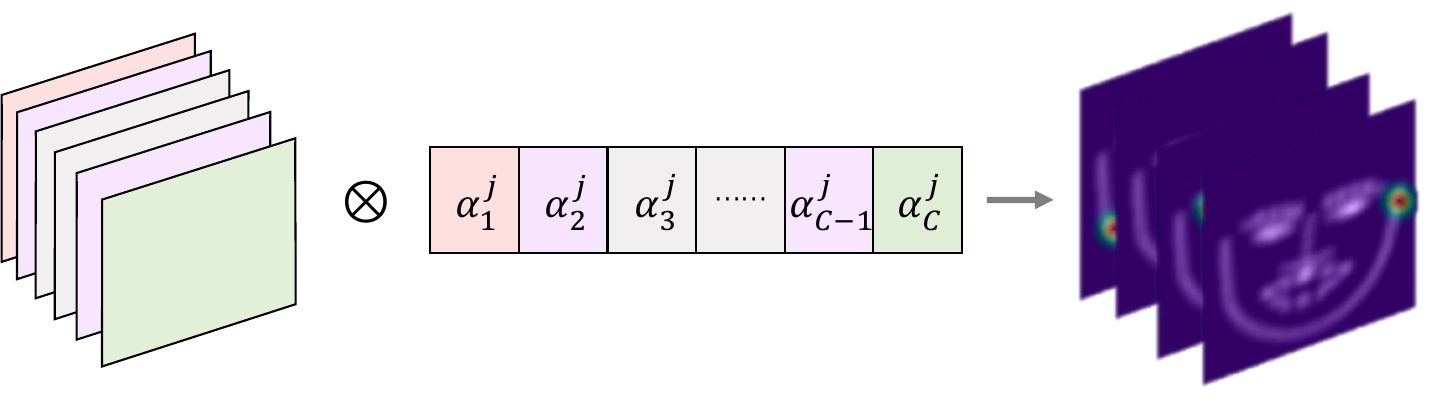}
    \caption{Heatmap is linear combination of channel-split features. $\bigotimes$ denotes dot-product. Intermediate supervision uses Conv2D with $1\times1$ kernel to convert the features extracted from a prediction block into heatmaps. 
    This implies channel-split features linearly expand the heatmap space. 
    Based on this observation, we take advantage of the self-attention mechanism to learn underlying geometric relations among landmarks via channel-split features.}
    \label{fig:linear_combination}
\end{figure}

The design of the channel-split ViT is based on the insight that the channel dimension of feature maps essentially represents the bases of the heatmap space.
Specifically, in the architecture, the prediction block outputs a feature map $F \in \mathbb{R}^{C\times H \times W}$ for intermediate supervision, where $C, H$ and $W$ denote the number of channels, height, and width. Then, the feature map $F$ can be split along the channels into $F=(f_1,f_2,...,f_C)$, where $f_m \in \mathbb{R}^{H \times W}$. Therefore, when considering using convolution operations to regress the feature map $F$ into the heatmaps, the heatmap for each landmark in the intermediate supervision is actually a linear combination of $f_m$:\\
\begin{equation}\label{eq:linear_combination}
    h_i = \sum^C_{m=1}{\alpha^i_m f_m}, \quad  (i=1,2,...,M), 
\end{equation}
where $M$ is the number of landmarks, $h_i$ is the heatmap of $i$-th landmark, and $\alpha^i_m$ is learnable parameters. 
In experiment, such linear combination can be implemented using a Conv2D layer with $1\times1$ kernel. Besides, ~\cref{fig:linear_combination} presents a visual explanation.

~\cref{eq:linear_combination} and~\cref{fig:linear_combination} show that the separated sub feature maps ($f_1$, $f_2$, ..., $f_C$) are actually linear bases of the heatmap space.
Since the heatmaps determine the coordinates of the landmarks, we can capture the inherent geometric relations among landmarks by analyzing the relations among the linear bases of heatmap space.    

In our implementation, we utilize transformer to capture the inherent relationships among the linear bases ($f_1$, $f_2$, ..., $f_C$), allowing for adaptive computation of their interconnections through the multi-head self-attention mechanism, which is the core component of the transformer architecture~\cite{NIPS2017_3f5ee243}. For self-containedness, we briefly describe the self-attention mechanism.  It projects an input sequence $X \in \mathbb{R}^{L\times d}$ into query $Q \in \mathbb{R}^{L\times d}$, key $K \in \mathbb{R}^{L\times d}$, and value $V \in \mathbb{R}^{L\times d}$ by three learnable matrices. The attention mechanism is formulated as follows:
\begin{equation}\label{eq:attention}
    Attention(Q,K,V) = softmax(\frac{QK^T}{\sqrt{d}})V.
\end{equation}
Then, our D-ViT is formally defined as:
\begin{equation}
    D\text{-}ViT = Conv\big(ViT(Pat)|| ViT(Chn)\big),
\end{equation}
where $Pat$ denotes the spatial-split patches, $Chn$ represents the channel-split features, $||$ refers to concatenation along the channel dimension and $Conv(\cdot)$ is a residual convolution block. Beneficial from such design, our D-ViT can leverage both spatial-split patches and channel-split features to extract image features and establish inherent relations among landmarks, thereby achieving new state-of-the-art results across three benchmarks. In the following discussion, we refer to the standard ViT with spatially split patches as spatial-split ViT, and the ViT with channel-split features as channel-split ViT.


\subsection{Long Skip Connection}
\label{Subsec:LSP}
Following the design paradigm of the widely used hourglasses networks~\cite{zhang2014facial,Newell2016StackedHN}, which often serve as backbones for facial landmark detection~\cite{Newell2016StackedHN,DBLP:journals/corr/abs-2104-03100,Wang_2019_ICCV,huang2021adnet,Zhou_2023_CVPR}, we repeat the prediction block in conjunction with intermediate supervision to consolidate feature processing. However, we found that when using residual connections between two sequential prediction blocks (\ie, ResCBSP in~\cref{fig:res_conn:res}), the detection performance is instead diminished by deeper prediction blocks. More details can be found in the ablation study in~\cref{Subsec:Ablation}.

The reason of this behavior is the supervision of outputs from intermediate prediction blocks. Specifically, during training, supervision at all intermediate stages compels the network to extract relevant information for estimating landmarks. However, this can also lead to the loss of some information; since shallow prediction blocks may not perform optimally, they might inadvertently discard useful information that could be better processed by deeper prediction blocks. 

Therefore, unlike previous methods that use residual connections between two consecutive prediction blocks, we propose using long skip connections (LSC) to distribute low-level image features, thereby making deeper network architectures feasible. Specifically, the LSC (shown as the upper black line in~\cref{fig:model}) originates from the low-level image features extracted by ResNet and transmits these features to each prediction block. As a result, each intermediate prediction block receives features extracted from the previous block as well as the complete low-level features. 

\subsection{Training Loss}
\label{Subsec:Loss}
We adopt the widely used soft-argmax operator to decode heatmaps into landmark positions. Let $h_i^j$ denote the heatmap for the $i$-th landmark predicted by $j$-th intermediate supervision. We denote the $k$-th pixel position in the heatmap as $g_k$ and the heatmap value at $g_k$ as $h_{ik}^j$. The corresponding landmark location for heatmap $h_i^j$ is given by: 
\begin{equation}
\centering
    \mu_i^j = \sum_{k=1}^{H \times W}{h_{ik}^j g_k}.
\end{equation}

The loss for the $j$-th intermediate supervision consists of two components: one for supervising landmark coordinates and the other for supervising the heatmaps, as shown in the following formula:
\begin{equation}
\label{eq:int_loss}
    \mathcal{L}_j =\sum_i \big(d_1(\mu^j_i, y_i) + \beta~ d_2(h^j_i, z_i)\big),
\end{equation}
where $y_i$ is the ground truth location of the $i$-th landmark, and $z_i$ is the corresponding heatmap defined by a Gaussian kernel. $\beta$ is a balance weight between coordinate and heatmap regression. $d_1$ and $d_2$ are loss functions for regressing coordinates and heatmaps. In this paper, we utilize smooth-L1 as $d_1$ and awing loss~\cite{Wang_2019_ICCV} as $d_2$. 

The total loss for optimizing our network is a combination of the loss terms from each intermediate supervision:
\begin{equation}\label{eq:total_loss}
\mathcal{L}_{total} = \sum^B_{j=1}{w^{j - B}}\mathcal{L}_j,
\end{equation}
where $B$ is the number of prediction blocks, and $w (w \geq 1)$ is an expanding factor that balances intermediate supervisions across different block outputs.

\section{Experiments}
In this section, we first introduce the experimental setup, including the datasets, evaluation metrics, and the implementation details in Section~\ref{Subsec:Exp:Setup}. Then, we compare our approach with state-of-the-art face landmark detection methods in Section~\ref{Subsec:Analysis:of:Results}. Finally, we perform ablation studies to analyze the design of our framework in Section~\ref{Subsec:Ablation}.

\subsection{Experimental Setup}\label{Subsec:Exp:Setup}
\textbf{Datasets.} For experimental evaluation, we consider three widely used datasets: \textbf{WFLW}~\cite{Wu_2018_CVPR}, \textbf{COFW}~\cite{Burgos-Artizzu_2013_ICCV}, and \textbf{300W}~\cite{Sagonas_2013_ICCV_Workshops}. Additionally, in the ablation experiments, we also employed a video-based dataset \textbf{WFLW-V}~\cite{Micaelli_2023_CVPR} for cross-dataset validation. The WFLW dataset is based on the WIDER dataset~\cite{Yang_2016_CVPR} and includes 7,500 training images and 2,500 test images, each with 98 labeled keypoints. The test set is further divided into six subsets: large-pose, expression, illumination, makeup, occlusion and blur, to assess algorithm performance under various conditions. For this dataset, we use the pre-cropped WFLW dataset from \mbox{Lan~\etal}~\cite{DBLP:journals/corr/abs-2104-03100} in our experiments. The COFW dataset features heavy occlusions and a wide range of head poses, comprising 1,345 training images and 507 test images. Each face is annotated with 29 landmarks. 300W dataset is a widely adopted dataset for face alignment and contains 3,148 images for training and 689 images for testing. The test set is divided into two subsets: common and challenge. All images are labeled with 68 landmarks. The \mbox{WFLW-V} dataset provides 1,000 videos, which are equally categorized into easy and hard subsets. In the cross-dataset validation, we train our model on the WFLW dataset and test it on every video frame from WFLW-V dataset.

\textbf{Evaluation Metrics.} For quantitative evaluation, three commonly used metrics are adopted: Normalized Mean Error (NME), Failure Rate (FR), and Area Under Curve (AUC). To calculate NME, we use the interocular distance for normalization in the WFLW and 300W datasets, and the interpupils distance for normalization in the COFW dataset. Same with previous works like STAR~\cite{Zhou_2023_CVPR} and LDEQ~\cite{Micaelli_2023_CVPR}, we report FR and AUC for the WFLW dataset with the cut-off threshold $10\%$.  For NME and FR, lower values indicate better performance, while for AUC, a higher value is preferable.

\begin{figure}[htbp]
  \centering
    \begin{subfigure}{0.95\linewidth}
        \includegraphics[width=\linewidth]{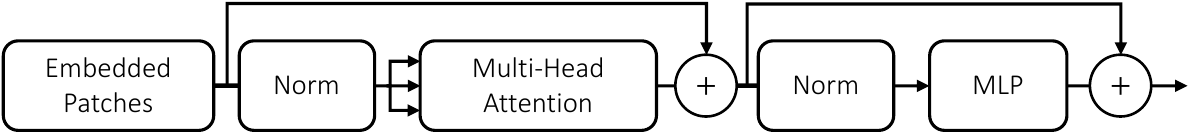}
        \caption{The architecture of ViT used in experiments.}\label{fig:vit}
        \vspace{1mm}
    \end{subfigure}

    \begin{subfigure}{0.95\linewidth}
        \includegraphics[width=\linewidth]{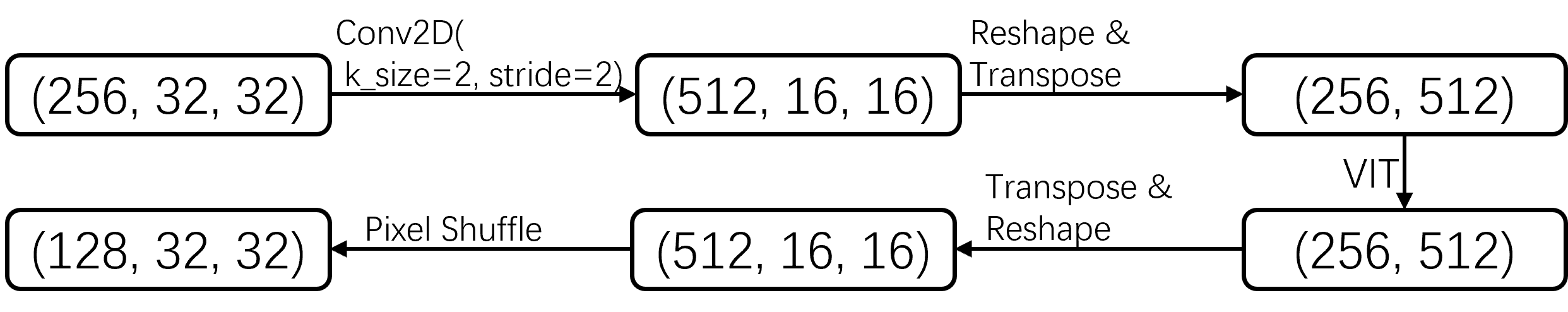}
        \vspace{-5mm}
        \caption{The architecture of  spatial-split ViT.}\label{fig:spatial-vit}
        \vspace{1mm}
    \end{subfigure}

    \begin{subfigure}{0.95\linewidth}
        \includegraphics[width=\linewidth]{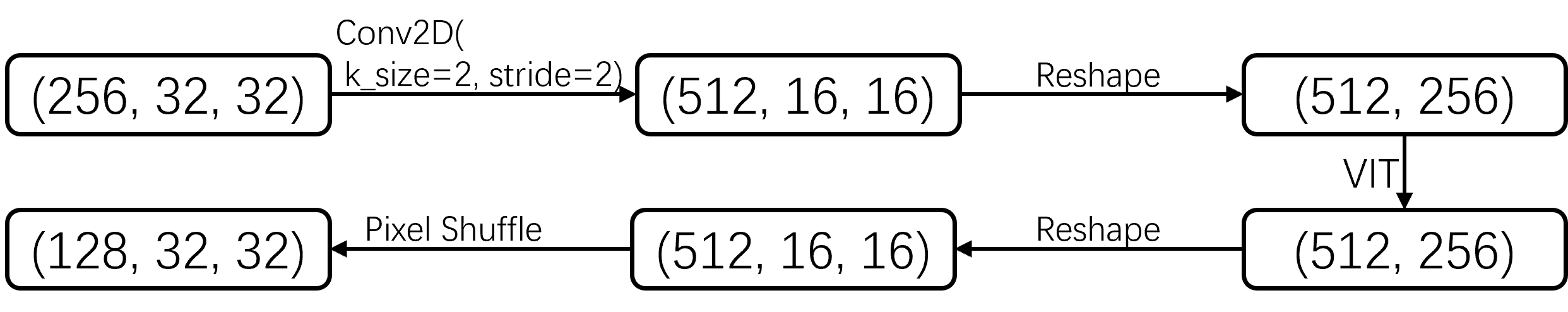}
        \vspace{-5mm}
        \caption{The architecture of channel-split ViT.}\label{fig:channel-vit}
        \vspace{2mm}
    \end{subfigure}

    \begin{subfigure}{0.95\linewidth}
        \includegraphics[width=\linewidth]{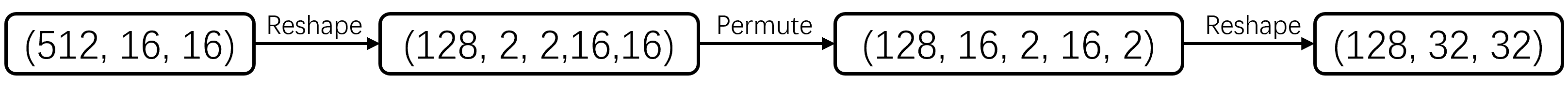}
        \caption{Pixel Shuffle.}\label{fig:pixelshuffle}
    \end{subfigure}\vspace{-2mm}

    \caption{Implementation details of our D-ViT.}
    \label{fig:impl_detail}
    \vspace{-0.2cm}
\end{figure}

\textbf{Implementation Details.} For the images fed into our network, the face regions are cropped and resized to $256\times 256$. Common image augmentations have been applied including random rotation, random translation, random occlusion, random blur, random color jitter, random horizontal flip, and random grayscale conversion. In our network, ResNet~\cite{He_2016_CVPR} is utilized to extract low-level image features. The ResNet in our model is built with bottleneck structure, and its number of parameters is about $13\%$ of the standard ResNet50. We stack 8 prediction blocks to sequentially predict $256 \times 32 \times 32$ feature maps for intermediate supervision. 
In spatial-split ViT, we apply a Conv2d layer for patch embedding. In channel-split ViT, we use another Conv2d layer to halve the spatial size to save memory. At the end of channel-split or spatial-split ViT, pixel shuffle~\cite{7780576} is used to restore the spatial dimensions to $32 \times 32$. More implementation details about D-ViT are shown in~\cref{fig:impl_detail}.
Ground-truth heatmaps are generated by the 2-dimensional Gaussian distribution with small variance~\cite{Zhang_2020_CVPR}.  We employ the Adam optimizer with initial learning rate $1\times 10^{-4}$. The learning rate is reduced by half for every 200 epochs, and we optimize the network parameters for a totoal of 500 epochs. The model is trained on two GPUs (Nvidia V100 16G), with a batch size of 16 per GPU.

\subsection{Comparison on Detection Accuracy}
\label{Subsec:Analysis:of:Results}
\begin{table*}[htpb]
\centering
\setlength{\extrarowheight}{0.9pt} 
\begin{tabular}{llccccccccccc}
\toprule
\multicolumn{2}{c}{\multirow{2}{*}{\textbf{Method}}} & \multicolumn{7}{c}{\textbf{WFLW}} & \multicolumn{1}{c}{\textbf{COFW}} & \multicolumn{3}{c}{\textbf{300W}} \\ \cmidrule(lr){3-9} \cmidrule(lr){10-10} \cmidrule(lr){11-13}
\multicolumn{1}{c}{} & \multicolumn{1}{c}{} & Full          & Pose   & Expr.   & Illum.  & Makeup        & Occl.     & Blur          & Full & Full & Comm. & Chal. \\ 
\cmidrule[\heavyrulewidth]{1-13}
LAB~\cite{Wu_2018_CVPR}&\scriptsize{CVPR'18}                     & 5.27          & 10.24         & 5.51          & 5.23          & 5.15          & 6.79          & 6.32          & -    & 3.49 & 2.98   & 5.19      \\
Wing ~\cite{Feng_2018_CVPR}&\scriptsize{CVPR'18}                   & 4.99          & 8.43          & 5.21          & 4.88          & 5.26          & 6.21          & 5.81          & 5.44 & -    & -      & -         \\
DecaFA~\cite{Dapogny_2019_ICCV}&\scriptsize{ICCV'19}                  & 4.62          & 8.11          & 4.65          & 4.41          & 4.63          & 5.74          & 5.38          & -    & 3.39 & 2.93   & 5.26      \\
HRNet~\cite{sun2019deep}&\scriptsize{CVPR'19}                   & 4.60          & 7.94          & 4.85          & 4.55          & 4.29          & 5.44          & 5.42          & -    & 3.32 & 2.87   & 5.15      \\
AS~\cite{qian2019aggregation}&\scriptsize{ICCV'19}                 & 4.39          & 8.42          & 4.68          & 4.24          & 4.37          & 5.60          & 4.86          & -    & 3.86 & 3.21   & 6.46      \\
LUVLI~\cite{kumar2020luvli}&\scriptsize{CVPR'20}                   & 4.37          & 7.56          & 4.77          & 4.30          & 4.33          & 5.29          & 4.94          & -    & 3.23 & 2.76   & 5.16      \\
AWing~\cite{Wang_2019_ICCV}&\scriptsize{ICCV'19}                   & 4.36          & 7.38          & 4.58          & 4.32          & 4.27          & 5.19          & 4.96          & 4.94 & 3.07 & 2.72   & 4.52      \\
SDL~\cite{Li2020StructuredLD}&\scriptsize{ECCV'20}                     & 4.21          & 7.36          & 4.49          & 4.12          & 4.05          & 4.98          & 4.82          & -    & 3.04 & 2.62   & 4.77      \\
ADNet~\cite{huang2021adnet}&\scriptsize{ICCV'21}                   & 4.14          & 6.96          & 4.38          & 4.09          & 4.05          & 5.06          & 4.79          & 4.68 & 2.93 & 2.53   & 4.58      \\
SLPT~\cite{Xia_2022_CVPR}&\scriptsize{CVPR'22}                    & 4.12          & 6.99          & 4.37          & 4.02          & 4.03          & 5.01          & 4.79          & 4.79 & 3.17 & 2.75   & 4.90      \\
HIH~\cite{DBLP:journals/corr/abs-2104-03100}&\scriptsize{ICCVW'21}                     & 4.08          & 6.87          & 4.06          & 4.34          & 3.85          & 4.85          & 4.66          & 4.63 & 3.09 & 2.65   & 4.89      \\
DTLD~\cite{9878969}&\scriptsize{CVPR'22}                    & 4.08          & -             & -             & -             & -             & -             & -             & -    & 2.96 & 2.59   & 4.50     \\
SPIGA~\cite{Prados-Torreblanca_2022_BMVC}&\scriptsize{BMVC'22}                    & 4.06          & 7.14          & 4.46        & \textcolor{blue}{4.00}          & 3.81         & 4.95   & 4.65         & -   & - & -   &  -     \\
STAR~\cite{Zhou_2023_CVPR}&\scriptsize{CVPR'23}                & 4.02          & \textcolor{blue}{6.79}           & 4.27          & \textcolor{red}{3.97}          &  3.84          & 4.80         & \textcolor{blue}{4.58}         & \textcolor{blue}{4.62} & \textcolor{blue}{2.87} & \textcolor{blue}{2.52}   & \textcolor{red}{4.32}      \\
LDEQ~\cite{Micaelli_2023_CVPR}&\scriptsize{CVPR'23}                    & \textcolor{blue}{3.92}          & 6.86          & \textcolor{blue}{3.94}          & 4.17          & \textcolor{blue}{3.75}          & \textcolor{blue}{4.77}          & 4.59          & -    & -    & -      & -         \\
FRA~\cite{gao2024self}&\scriptsize{CVPR'24} & 4.11 & - & - & - & - & - & - & - & 2.91 & 2.60 & \textcolor{blue}{4.46} \\
\midrule
\textbf{Ours}&-                    & \textcolor{red}{3.75} & \textcolor{red}{6.43} & \textcolor{red}{3.85} & 4.06 & \textcolor{red}{3.57} & \textcolor{red}{4.47} & \textcolor{red}{4.37} & \textcolor{red}{4.13} & \textcolor{red}{2.85} & \textcolor{red}{2.43} &  4.56  \\
\bottomrule
\end{tabular}
\caption{
Quantitative comparison with previous state-of-the-art methods on three public datasets using the NME metric.
The \textcolor{red}{best} and \textcolor{blue}{second} best results are marked in colors of red and blue, respectively. It can be seen that our method achieves new SOTA results on the full test sets of all three datasets.
}\label{table:nme}
\end{table*}
\begin{table*}[htbp]
\centering
\setlength{\extrarowheight}{1pt} 
\begin{tabular}{lcccccccccccccc}
\toprule
\multicolumn{1}{c}{\multirow{2}{*}{\textbf{Method}}} & \multicolumn{7}{c}{\textbf{FR$_{10}(\downarrow)$}}                         & \multicolumn{7}{c}{\textbf{AUC$_{10}(\uparrow)$}}  \\ \cmidrule(lr){2-8} \cmidrule(lr){9-15}
\multicolumn{1}{c}{}                        & Full & Pose  & Exp. & Ill. & Mu.  & Occ.  & Blur  & Full & Pose & Exp. & Ill. & Mu.  & Occ. & Blur \\ \cmidrule[\heavyrulewidth]{1-15}
LAB~\cite{Wu_2018_CVPR}                                          & 7.56 & 28.83 & 6.37 & 6.73 & 7.77 & 13.72 & 10.74 & 53.2 & 23.5 & 49.5 & 54.3 & 53.9 & 44.9 & 46.3 \\
HRNet~\cite{sun2019deep}                                        & 4.64 & 23.01 & 3.5  & 4.72 & 2.43 & 8.29  & 6.34  & 52.4 & 25.1 & 51   & 53.3 & 54.5 & 45.9 & 45.2 \\
AS~\cite{qian2019aggregation}                                      & 4.08 & 18.1  & 4.46 & 2.72 & 4.37 & 7.74  & 4.4   & 59.1 & 31.1 & 54.9 & 60.9 & 58.1 & 51.6 & 55.1 \\
LUVLi~\cite{kumar2020luvli}                                        & 3.12 & 15.95 & 3.18 & 2.15 & 3.4  & 6.39  & 3.23  & 55.7 & 31   & 54.9 & 58.4 & 58.8 & 50.5 & 52.5 \\
AWing~\cite{Wang_2019_ICCV}                                        & 2.84 & 13.5  & 2.23 & 2.58 & 2.91 & 5.98  & 3.75  & 57.2 & 31.2 & 51.5 & 57.8 & 57.2 & 50.2 & 51.2 \\
SDL~\cite{Li2020StructuredLD}                                          & 3.04 & 15.95 & 2.86 & 2.72 & \textcolor{blue}{1.45} & 5.29  & 4.01  & 58.9 & 31.5 & 56.6 & 59.5 & 60.4 & 52.4 & 53.3 \\
ADNet~\cite{huang2021adnet}                                        & 2.72 & 12.72 & 2.15 & 2.44 & 1.94 & 5.79  & 3.54  & 60.2 & 34.4 & 52.3 & 58   & 60.1 & 53   & 54.8 \\
SLPT~\cite{Xia_2022_CVPR}                                         & 2.72 & 11.96 & 1.59 & 2.15 & 1.94 & 5.7   & 3.88  & 59.6 & 34.9 & 57.3 & 60.3 & 60.8 & 52   & 53.7 \\
HIH~\cite{DBLP:journals/corr/abs-2104-03100}                                          & 2.6  & 12.88 & 1.27 & 2.43 & \textcolor{blue}{1.45} & 5.16  & 3.1   & 60.5 & 35.8 & 60.1 & 61.3 & 61.8 & 53.9 & 56.1 \\
FRA~\cite{gao2024self} & 2.53 & - & - & - & - & -  & -  & 59.1 & - & - & - & - & - & - \\
SPIGA~\cite{Prados-Torreblanca_2022_BMVC}                                        & \textcolor{blue}{2.08} & \textcolor{blue}{11.66} & 2.23 & \textcolor{blue}{1.58} & 1.46 & \textcolor{blue}{4.48}  & \textcolor{blue}{2.20}  & 60.6 & 35.3 & 58   & 61.3 & 62.2 & 53.3 & 55.3 \\
STAR~\cite{Zhou_2023_CVPR}                                     & 2.32 & 11.69 & 2.24 & \textcolor{blue}{1.58} & \textcolor{red}{0.98} & 4.76  & 3.24  & 60.5 & 36.2 & 58.4 & 60.9 & 62.2 & 53.8 & 55.1 \\
LDEQ~\cite{Micaelli_2023_CVPR}                                         & 2.48 & 12.58 & \textcolor{blue}{1.59} & 2.29 & 1.94 & 5.36  & 2.84  & \textcolor{blue}{62.4} & \textcolor{blue}{37.3} & \textcolor{blue}{61.4} & \textcolor{blue}{63.1} & \textcolor{blue}{63.1} & \textcolor{blue}{55.2} & \textcolor{blue}{57.4} \\ 
\midrule
\textbf{Ours}                                         & \textcolor{red}{1.76} & \textcolor{red}{8.28}  & \textcolor{red}{1.27} & \textcolor{red}{1.29} & 1.94 & \textcolor{red}{3.80}  & \textcolor{red}{2.07}  & \textcolor{red}{63.7} & \textcolor{red}{40.1} & \textcolor{red}{62.6} & \textcolor{red}{64.7} & \textcolor{red}{64.7} & \textcolor{red}{57.1} & \textcolor{red}{58.6} \\ \bottomrule
\end{tabular}
\caption{FR$_{10}$ and AUC$_{10}$ on the WFLW test set. The \textcolor{red}{best} and \textcolor{blue}{second} best results are marked in colors of red and blue, respectively. The results demonstrate the robustness and effectiveness of our proposed model.}
\vspace{-0.3cm}
\label{table:fr_auc}
\end{table*}
In this section, we compare our method with previous state-of-the-art baselines. 
The NME across three datasets are reported in~\cref{table:nme}, while the FR and AUC for WFLW dataset are shown in~\cref{table:fr_auc}. The results presented in both tables demonstrate that our approach surpasses the previous baselines and achieves a new SOTA across all three datasets.

Compared to other transformer-based methods~\cite{9878969, Xia_2022_CVPR}, our approach achieves an improvement of 0.33 and 0.37 in NME on the full WFLW dataset over DTLD~\cite{9878969} and SLPT~\cite{Xia_2022_CVPR}, respectively.
This demonstrates that our proposed D-ViT and long skip connection have a positive impact on the performance of transformers. 
Additionally, our proposal also outperforms recent CNN-based methods~\cite{Zhou_2023_CVPR,Micaelli_2023_CVPR} that achieved state-of-the-art results.
Specifically, our NME score is 0.27 and 0.17 better than that of STAR~\cite{Zhou_2023_CVPR} and LDEQ~\cite{Micaelli_2023_CVPR} on the full WFLW test set. 
Furthermore, for the Pose and Occlusion subsets, which contain severe occlusions, our method significantly improves the detection performance, demonstrating that our network is capable of capturing the contextual features of images and the intrinsic relationships between landmarks.
\begin{figure*}[!t]
\centering
    \captionsetup[subfigure]{labelformat=empty}
    \begin{subfigure}{0.16\linewidth}
        \includegraphics[width=\linewidth]{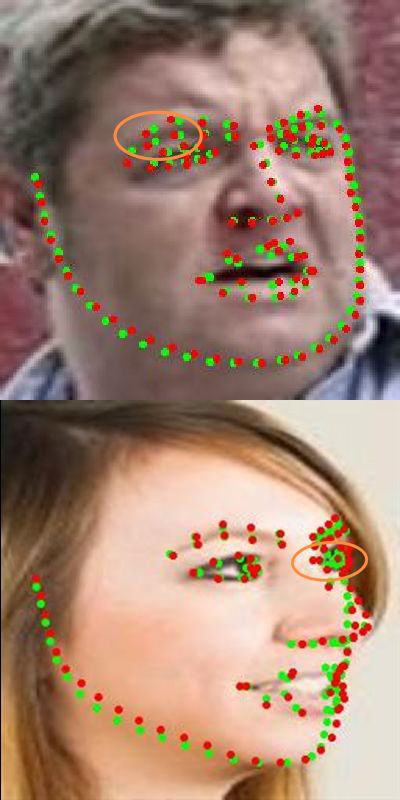}
        \caption{Spatial-split}
    \end{subfigure}\hspace{-1mm}
    \begin{subfigure}{0.16\linewidth}
        \includegraphics[width=\linewidth]{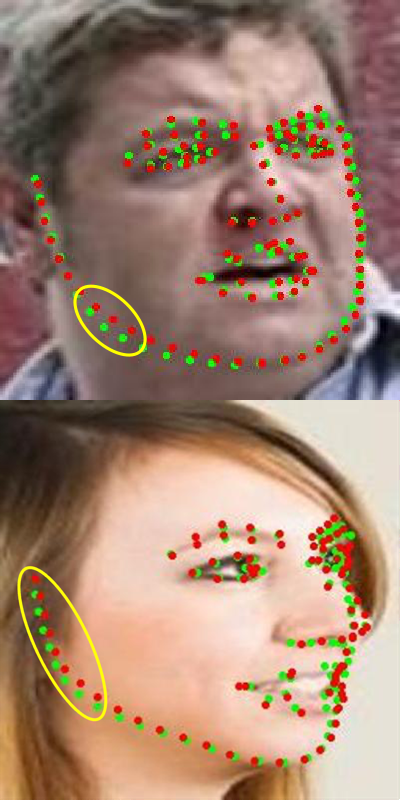}
        \caption{Channel-split}
    \end{subfigure}\hspace{-1mm}
    \begin{subfigure}{0.16\linewidth}
        \includegraphics[width=\linewidth]{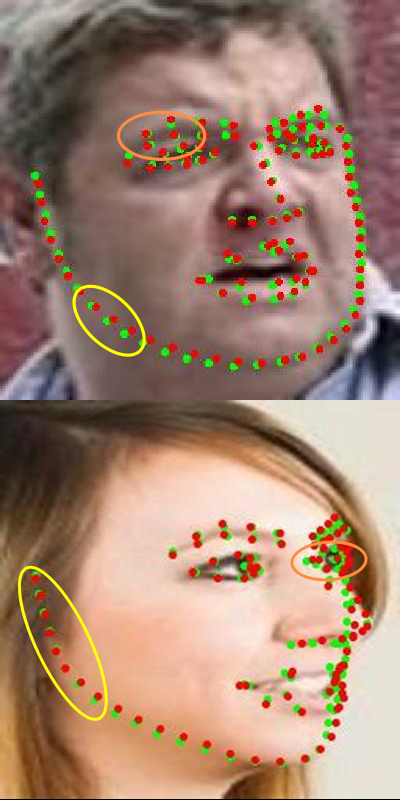}
        \caption{D-ViT}
    \end{subfigure}
    \begin{subfigure}{0.16\linewidth}
        \includegraphics[width=\linewidth]{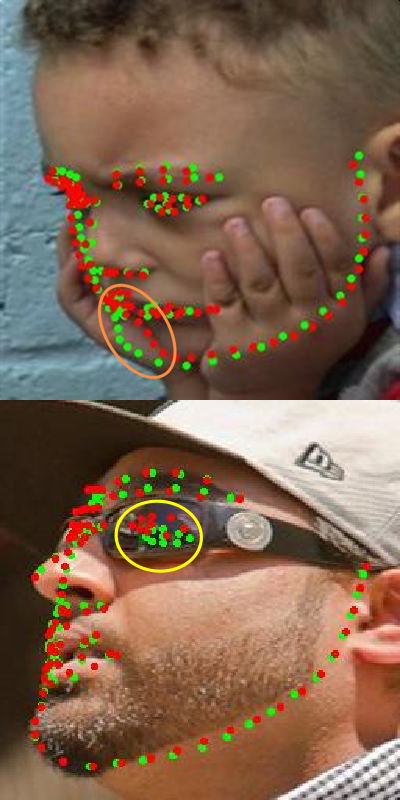}
        \caption{Spatial-split}
    \end{subfigure}\hspace{-1mm}
    \begin{subfigure}{0.16\linewidth}
        \includegraphics[width=\linewidth]{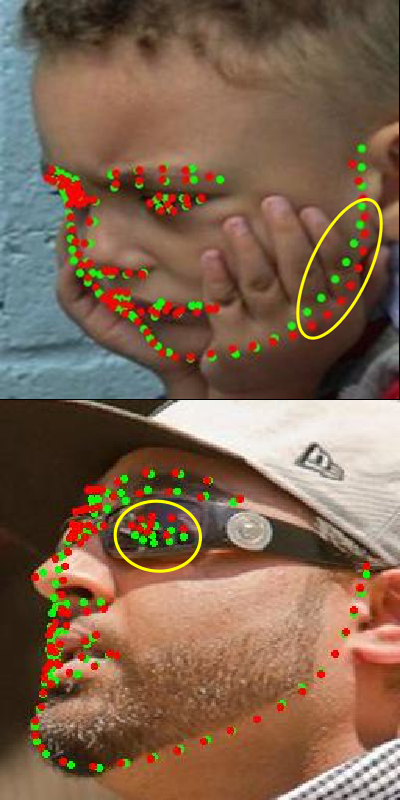}
        \caption{Channel-split}
    \end{subfigure}\hspace{-1mm}
    \begin{subfigure}{0.16\linewidth}
        \includegraphics[width=\linewidth]{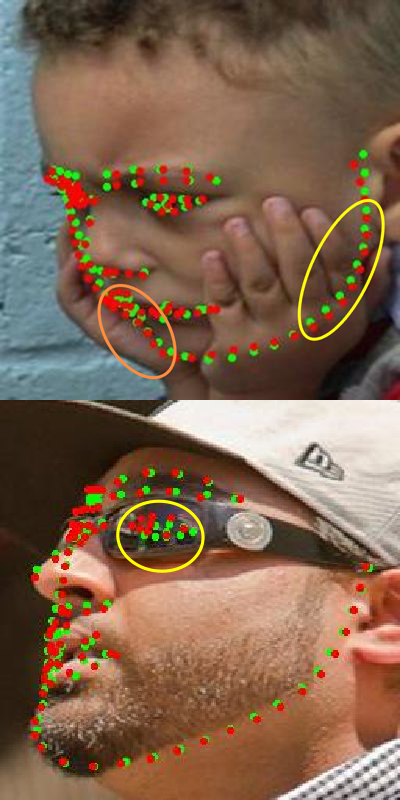}
        \caption{D-ViT}
    \end{subfigure}
    \vspace{-0.2cm}
    \caption{Qualitative results of different prediction blocks on WFLW dataset. Green and red points represent the predicted and ground-truth landmarks, respectively. Orange or Yellow circles indicate the clear failures, which can be improved with help of D-ViT.}
    \label{fig:dvit_comparison}
\end{figure*}
\begin{table*}[htbp]
\centering
\setlength{\tabcolsep}{2mm}{
\begin{tabular}{ccccccc}
    \toprule
     \multicolumn{1}{c}{\multirow{2}{*}{\textbf{Dataset}}} & \multicolumn{3}{c}{\textbf{Connection Strategy}} & \multicolumn{3}{c}{\textbf{Block Name}}        \\ \cmidrule(lr){2-4} \cmidrule(lr){5-7}
     & ResCBSP       & DenC       & LSC         & Spatial-split & Channel-split & D-ViT \\ \cmidrule[\heavyrulewidth]{1-7}
COFW & 4.14           & 4.14         & \textbf{4.13}        & 4.18          & 4.20          & \textbf{4.13}  \\
300W & 2.89           & 2.90         & \textbf{2.85}       & 2.87          & 2.96          & \textbf{2.85}  \\ \bottomrule
\end{tabular}}
\vspace{-0.2cm}
\caption{Comparisons of different connection strategies and different prediction blocks on dataset COFW and 300W by using  8 prediction blocks. NME scores 
 are reported.}
 \label{table:comparison_cofw_300w}
 \vspace{-0.2cm}
\end{table*}


By exploring the relationships between bases in the heatmap space to model the underlying geometric relations among landmarks, our method achieves a significant improvement of 0.49 in NME on the COFW dataset, which contains heavy occlusions and a wide range of head poses, compared to the previous SOTA baseline, STAR~\cite{Zhou_2023_CVPR}. Moreover, our approach also achieves the lowest NME on the full 300W test set.

The FR$_{10}$ and AUC$_{10}$ scores on the WFLW dataset, as reported in~\cref{table:fr_auc}, demonstrate the robustness and effectiveness of our proposed model. Specifically, our method outperforms previous state-of-the-art methods~\cite{Prados-Torreblanca_2022_BMVC,Micaelli_2023_CVPR} by 0.32 and 1.30 in FR$_{10}$ and AUC$_{10}$, respectively.

\subsection{Ablation Studies}\label{Subsec:Ablation}
In this section, various ablation studies are conducted to verify the specific design decisions in our model architecture. Discussion about the selections of hyper parameters is also included.

\begin{figure*}[htbp]
    \begin{subfigure}{0.47\linewidth}
        \includegraphics[width=\linewidth]{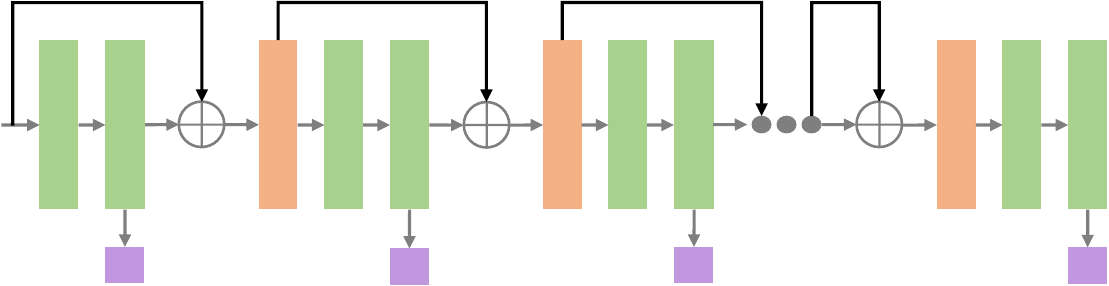}
        \caption{ResCBSP}\label{fig:res_conn:res}
    \end{subfigure}\hspace{\fill}
    \begin{subfigure}{0.47\linewidth}
        \includegraphics[width=\linewidth]{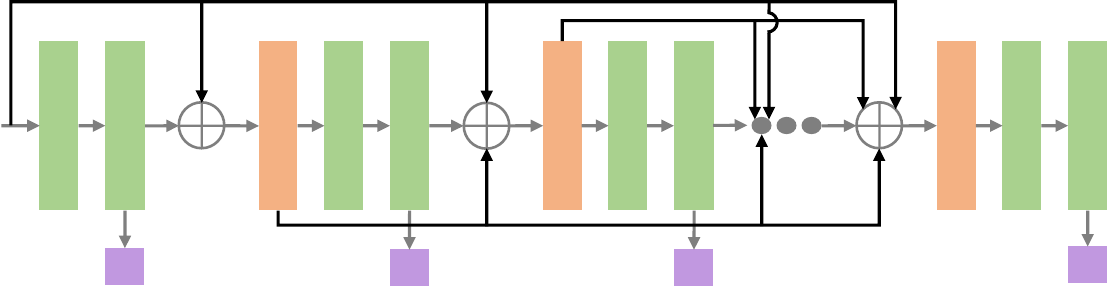}
        \caption{DenC}\label{fig:res_conn:dense}
    \end{subfigure}
    \vspace{-0.2cm}
    \caption{Illustrations of alternative skip connections. ResCBSP denotes the residual connections between two sequential prediction blocks. DenC denotes dense connections where any two prediction blocks have a skip connection. Refer to ~\cref{fig:model} for definition of prediction block and colorful rectangles.}
    \label{fig:res_conn}
    \vspace{-0.4cm}
\end{figure*}

\textbf{D-ViT.} 
Our Dual Vision Transformer (D-ViT) utilizes two types of Vision Transformers (ViTs), specifically the spatial-split ViT and channel-split ViT, to extract spatial features from images and underlying geometric features among landmarks, respectively.
To demonstrate the necessity of incorporating the channel-split ViT for exploring the relationships between heatmap bases, we report in~\cref{table:dvit_comparison} the performance on the WFLW dataset when using spatial-split ViT, channel-split ViT, and our proposed D-ViT separately to construct the prediction blocks.
Additionally,~\cref{table:comparison_cofw_300w} presents the NME performance of different prediction blocks on the COFW and 300W datasets.
From the two tables, it can be observed that by incorporating the less effective channel-split ViT to construct D-ViT actually leads to more accurate detection. This indicates that exploring relationships among heatmap bases has a positive effect on enhancing accurate predictions.
~\cref{fig:dvit_comparison} shows some qualitative visualizations. With the help of D-ViT, our model is able to accurately detect landmarks in various scenarios, such as occlusion, expression, blur, and large pose. 

\begin{figure}[htbp]
    \centering
    \includegraphics[width=\linewidth]{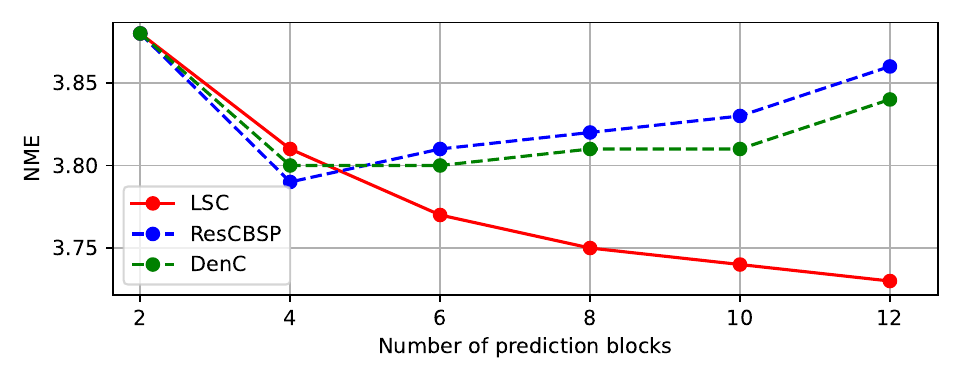}
    \vspace{-0.8cm}
    \caption{NME against number of prediction blocks on WFLW dataset. When the number is 2, three strategies essentially represent the same structure. When the number exceeds 4, the performance of ResCBSP or DenC starts to get worse. While with our proposed LSC, the model can benefit from a deeper architecture.}
    \label{fig:num_pred_blk}
    \vspace{-0.2cm}
\end{figure}
\begin{table}[htbp]
    \begin{subtable}[t]{0.49\textwidth}
    \centering
\begin{tabular}{cccc}
\toprule
Block Name    & NME$(\downarrow)$  & FR$_{10}(\downarrow)$ & AUC$_{10}(\uparrow)$ \\
\cmidrule[\heavyrulewidth]{1-4}
Spatial-Split & 3.82 &  2.08&   63.2  \\
Channel-Split & 3.87 &  2.36  &  63.2   \\
D-ViT         & \textbf{3.75} &  \textbf{1.76}  &  \textbf{63.7}  \\
\bottomrule
\end{tabular}
\caption{
Quantitative comparison of different prediction blocks.
}
\label{table:dvit_comparison}
        \vspace{0.2cm}
    \end{subtable}\hspace{5mm}
    \begin{subtable}[t]{0.49\textwidth}
    \centering
\begin{tabular}{cccc}
\toprule
Conn. Strategy & NME$(\downarrow)$& FR$_{10}(\downarrow))$ & AUC$_{10}(\uparrow)$ \\
\cmidrule[\heavyrulewidth]{1-4}
ResCBSP &3.82   &   2.16 &  63.2   \\
DenC &3.81   &   2.16 &  63.2   \\
LSC    &  \textbf{3.75}   & \textbf{1.76}   & \textbf{63.7}   \\
\bottomrule
\end{tabular}
\caption{Quantitative comparison of different connection strategies.}
\label{table:conn_comparison}
    \end{subtable}
    \caption{Comparison results of different prediction blocks (a) and different connection strategies (b) with 8 prediction blocks on WFLW dataset.}
\label{table:ablation_on_wflw}
\end{table} 
\textbf{Long Skip Connection.}
Connection strategies for prediction blocks are typically either residual connections between sequential predictions (ResCBSP)~\cite{he2016deep} or dense connections (DenC)~\cite{huang2017densely}, as illustrated in~\cref{fig:res_conn:res} and~\cref{fig:res_conn:dense}. Besides, residual connections are commonly used in stacked Hourglasses (HGs) networks~\cite{zhang2014facial,Newell2016StackedHN}, which often serve as backbones for facial landmark detection~\cite{Newell2016StackedHN,DBLP:journals/corr/abs-2104-03100,Wang_2019_ICCV,huang2021adnet,Zhou_2023_CVPR}. However,~\cref{fig:num_pred_blk} indicates that using these two strategies to connect prediction blocks built on ViTs results in diminished performance as the number of blocks increases. This is because intermediate supervision can lead to the loss of useful information. To address this, we propose using long skip connections (LSC) to distribute low-level image features from ResNet to each prediction block, thereby making deeper network architectures feasible.
Additionally, the quantitative comparison of different connection strategies with 8 prediction blocks reported in~\cref{table:conn_comparison} on the WFLW dataset, along with the results in~\cref{table:comparison_cofw_300w} on the COFW and 300W datasets, both demonstrate the superior effectiveness of our proposed LSC.


\begin{table}[htbp]
    \begin{subtable}[t]{0.495\textwidth}
    \centering
\setlength{\tabcolsep}{4mm}{
\begin{tabular}{ccc}
\toprule
Block Name    & Easy & Hard \\
\cmidrule[\heavyrulewidth]{1-3}
Spatial-split & 1.74 & 2.96 \\
Channel-split & 1.69 & 3.12 \\
D-ViT         & \textbf{1.66} & \textbf{2.91}\\
\bottomrule
\end{tabular}}
\caption{Quantitative comparison of different prediction blocks.}
\label{table:wflw_v_pred_blk}
        \vspace{0.2cm}
    \end{subtable}
    \begin{subtable}[t]{0.495\textwidth}
    \centering
\setlength{\tabcolsep}{3mm}{
\begin{tabular}{ccc}
\toprule
Connection Strategy  & Easy & Hard \\
\cmidrule[\heavyrulewidth]{1-3}
ResCBSP & 1.77 & 3.12 \\
DenC & 1.74 & 3.06 \\
LSC & \textbf{1.66} & \textbf{2.91} \\
\bottomrule
\end{tabular}}
\caption{Quantitative comparison of different connection strategies.}
\label{table:wflw_v_conn}
    \end{subtable}
    \vspace{-0.2cm}
    \caption{Cross-dataset Validations. We train all models on WFLW dataset, and report NME on the two subsets of WFLW-V dataset. The results further demonstrate the effectiveness of our design.}
    \vspace{-0.1cm}
    \label{table:cross}
\end{table}
\textbf{Cross-dataset Validations.} To further validate our design decisions, we conduct a cross-dataset validation experiment that includes quantitative comparisons of different prediction blocks and connection strategies, similar to the one described in~\cref{table:ablation_on_wflw}. However, different from that experiment, we train the models on the WFLW dataset and evaluate them on subsets of the WFLW-V dataset, \ie, the easy set and the hard set. 
The results reported in~\cref{table:cross} indicate that our proposed D-ViT and LSP still achieve the best NME score in the cross-dataset validations, demonstrating superior generalization.

\begin{table}[htbp]
\centering
\setlength{\tabcolsep}{3mm}{
\begin{tabular}{ccccc}
\toprule
$w$   & 1.0  & 1.2  & 1.4  & 1.6 \\ 
\midrule
NME$(\downarrow)$ & 3.78 & \textbf{3.75} & 3.77 & 3.77 \\
\bottomrule
\end{tabular}
}
\caption{Analysis of weight $w$ in \cref{eq:total_loss}. We report NME scores with varying  $w$ on WFLW dataset. 
}
\label{table:weights}
\vspace{-0.3cm}
\end{table}
\textbf{Weight for Intermediate Supervision.} Weight $w$ in~\cref{eq:total_loss} is a balance between multiple intermediate supervisions. To study the influence, we carried out experiments with $w$ ranging from 1.0 to 1.6, as shown in~\cref{table:weights}. 
Our model achieves the best performance with $w$ set to 1.2. Thus, we choose 1.2 as the default setting.
\section{Conclusion}
This paper introduces a new approach for facial landmark detection based on our proposed dual vision transformers, which extract image features through spatial-split features and learn inherent geometric relations through channel-split features. Extensive experiments demonstrate that D-ViT plays an effective role in facial landmark detection, achieving new state-of-the-art performance on three benchmarks. Additionally, various ablation studies are conducted to demonstrate the necessity of the design choices in our network. Moreover, we also investigate the effect of different connection strategies between prediction blocks, 
revealing that our proposed long skip connection allows the network to incorporate more prediction blocks to improve accuracy without losing useful features in deeper blocks.


\clearpage
{\small
\bibliographystyle{ieee_fullname}
\bibliography{egbib}
}

\end{document}


\title{Cascaded Dual Vision Transformer for Accurate Facial Landmark Detection}

\maketitle
\appendix

In this supplementary material, we provide more details and results omitted from the main paper for brevity. 
Specifically, in Sec.~\ref{Sec:GPU}, we introduce the GPU memory requirement; in Sec.~\ref{Sec:Capacity}, we compare our method by training and testing networks with similar computational capacity; in Sec.~\ref{Sec:Resolution}, we investigate the impact of input image resolution; and in Sec.~\ref{Sec:Visual}, we present visual comparisons on the COFW and 300W datasets.
\section{GPU Memory Requirement}
\label{Sec:GPU}
In~\cref{table:memory}, we report the memory required for each GPU during training, as well as the number of parameters for different numbers of prediction blocks.

\begin{table}[htbp]
\centering
\small
\setlength{\extrarowheight}{1pt} 
\begin{adjustbox}{trim=0pt 0pt 1mm 0pt, clip}
\begin{tabular}{ccccccc}
\toprule
\#Pred. Blocks & 2    & 4    & 6    & 8    & 10    & 12    \\
\midrule
Memory (GB)     & 4.4  & 6.3  & 8.2  & 10.3 & 12.1  & 14.2  \\
\#Param. (M)       & 24.4 & 48.4 & 72.4 & 96.4 & 120.4 & 144.4\\
\bottomrule
\end{tabular}
\end{adjustbox}
\caption{Memory required for each GPU during training, and number of parameters for different numbers of prediction blocks.}
\label{table:memory}
\end{table}
\section{Comparison on Similar Compute Capacity}
\label{Sec:Capacity}
Our proposed Long Skip Connection avoid losing useful information due to intermediate supervision and make deeper network architectures feasible.
However, improved performance is not solely attributed to in
creased computational capacity. When we use 4 prediction blocks and reduce the dimension of the feature maps to (160, 32, 32), the number of parameters in our network is comparable to other baselines. As reported in~\cref{tbl:param}, the NME score still surpasses the previous SOTA method LDEQ~\cite{Micaelli_2023_CVPR} by 0.08, indicating the effectiveness of our proposed architecture.

\begin{table}[htbp]
\centering
\small
\setlength{\extrarowheight}{1pt} 
\begin{adjustbox}{trim=0pt 0pt 1.2mm 0pt, clip}
\begin{tabular}{>{\raggedright}p{1.2cm}cccc}
\toprule
Method             & \#Param. (M) & NME$(\downarrow)$  & FR$_{10}(\downarrow)$   & AUC$_{10}(\uparrow)$   \\
\midrule
HIH~\cite{DBLP:journals/corr/abs-2104-03100}                & 22.7       & 4.08 & 2.60 & 60.5  \\
SPIGA~\cite{Prados-Torreblanca_2022_BMVC}              & 60.3          & 4.06 & \textcolor{red}{2.08} & 60.6  \\
STAR~\cite{Zhou_2023_CVPR}           & 13.4       & 4.02 & \textcolor{blue}{2.32} & 60.5  \\
LDEQ~\cite{Micaelli_2023_CVPR}               & 21.8       & \textcolor{blue}{3.92} & 2.48 & \textcolor{blue}{62.4}  \\
Ours\_nstack4 & 21.0       & \textcolor{red}{3.84} & 2.44 & \textcolor{red}{63.3} \\
\bottomrule
\end{tabular}
\end{adjustbox}
\caption{Comparisons on WFLW dataset. We reduce the number of network parameters to 21M, denoted as “Ours\_nstack4". Our proposed method still shows effectiveness.}\label{tbl:param}
\end{table}

\section{Comparison on Different Image Resolutions}
\label{Sec:Resolution}
\begin{figure}[htbp]
  \centering
  \includegraphics[width=0.9\linewidth]{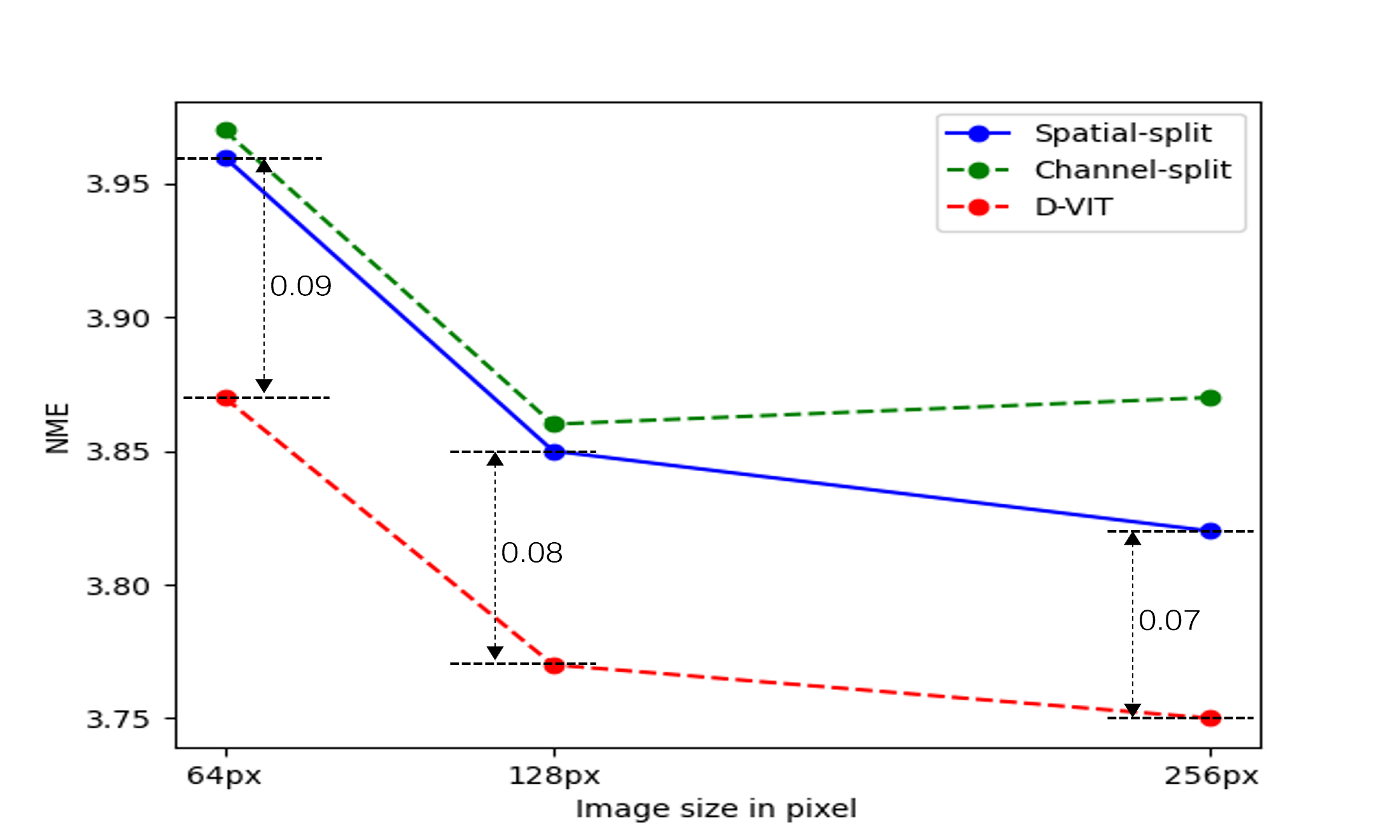}
\caption{NME against different input image sizes on WFLW dataset.}\label{fig:imagesize}
\end{figure}

We investigate the influence of different input image resolutions, as shown in~\cref{fig:imagesize}. Specifically, D-VIT improves the performance by 0.09, 0.08 and 0.07 at resolutions of 64px, 128px, and 256px, respectively, indicating that our proposed method is not sensitive to the input image size.
\section{Visual Results on COFW and 300W}
\label{Sec:Visual}
\begin{figure}[htbp]
\centering
    \captionsetup[subfigure]{labelformat=empty}
    \begin{subfigure}{0.33\linewidth}
        \includegraphics[width=\linewidth]{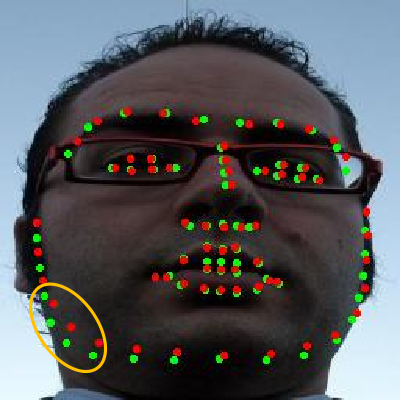}
    \end{subfigure}\hspace{-1mm}
    \begin{subfigure}{0.33\linewidth}
        \includegraphics[width=\linewidth]{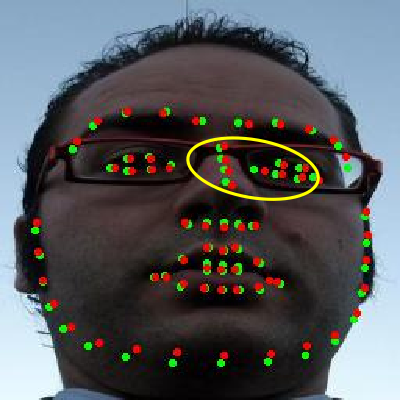}
    \end{subfigure}\hspace{-1mm}
    \begin{subfigure}{0.33\linewidth}
        \includegraphics[width=\linewidth]{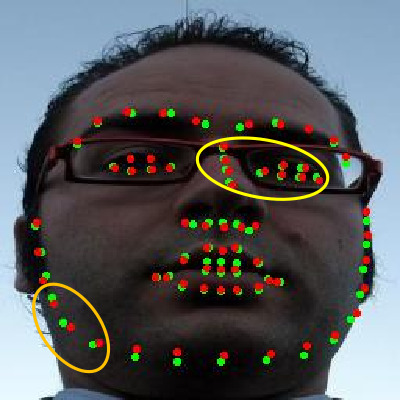}
    \end{subfigure}\vspace{-1mm}
    
    \begin{subfigure}{0.33\linewidth}
        \includegraphics[width=\linewidth]{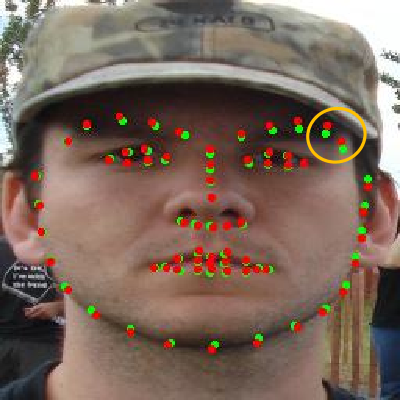}
        \caption{Spatial-split}
    \end{subfigure}\hspace{-1mm}
    \begin{subfigure}{0.33\linewidth}
        \includegraphics[width=\linewidth]{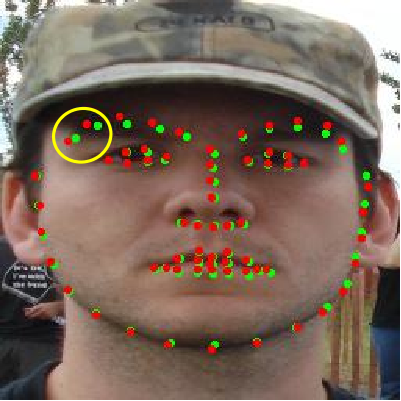}
        \caption{Channel-split}
    \end{subfigure}\hspace{-1mm}
    \begin{subfigure}{0.33\linewidth}
        \includegraphics[width=\linewidth]{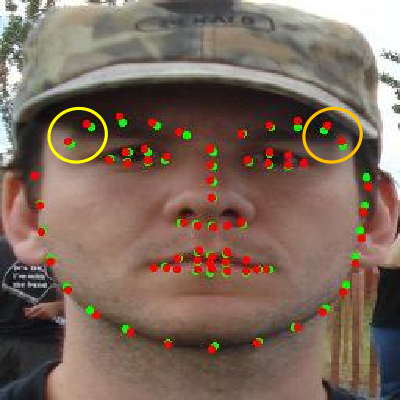}
        \caption{D-ViT}
    \end{subfigure}
    \caption{Visual comparison of different prediction blocks on 300W. Green and red points represent the predicted and ground-truth
landmarks, respectively.}\label{fig:300w:pred_blk}
\end{figure}

\begin{figure}[htbp]
\centering
    \captionsetup[subfigure]{labelformat=empty}
    \begin{subfigure}{0.33\linewidth}
        \includegraphics[width=\linewidth]{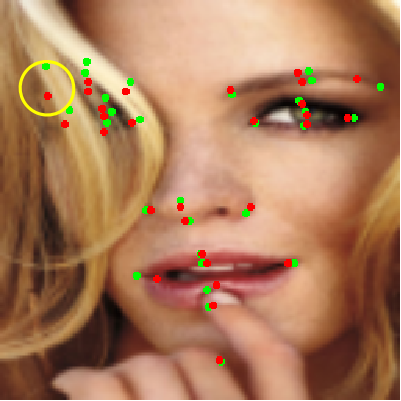}
    \end{subfigure}\hspace{-1mm}
    \begin{subfigure}{0.33\linewidth}
        \includegraphics[width=\linewidth]{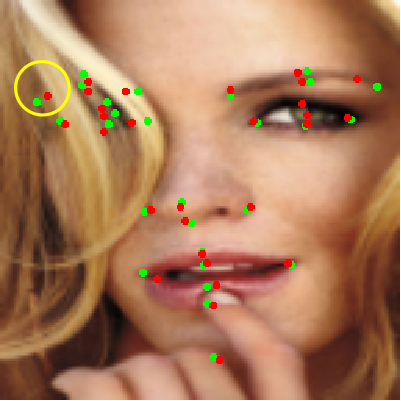}
    \end{subfigure}\hspace{-1mm}
    \begin{subfigure}{0.33\linewidth}
        \includegraphics[width=\linewidth]{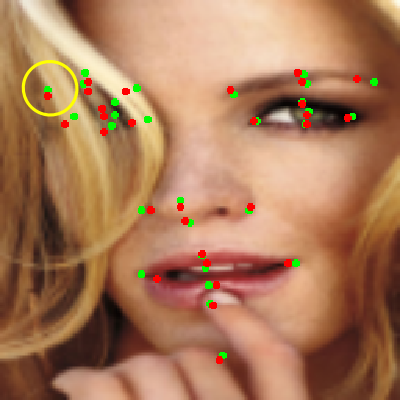}
    \end{subfigure}\vspace{-1mm}
    
    \begin{subfigure}{0.33\linewidth}
        \includegraphics[width=\linewidth]{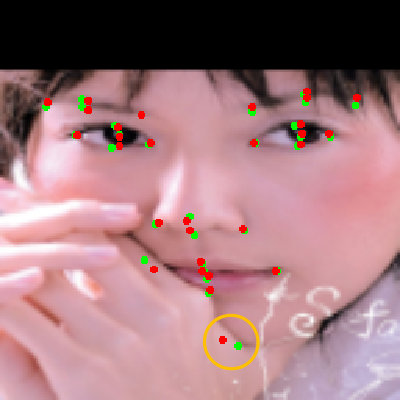}
        \caption{Spatial-split}
    \end{subfigure}\hspace{-1mm}
    \begin{subfigure}{0.33\linewidth}
        \includegraphics[width=\linewidth]{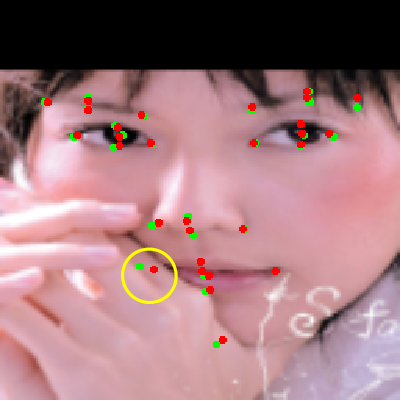}
        \caption{Channel-split}
    \end{subfigure}\hspace{-1mm}
    \begin{subfigure}{0.33\linewidth}
        \includegraphics[width=\linewidth]{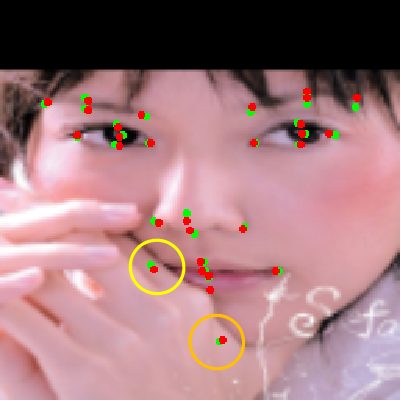}
        \caption{D-ViT}
    \end{subfigure}
    \caption{Visual results on the COFW dataset which contains heavy occlusions. 
    Geometric relations among landmarks play a crucial role in accurately predicting landmarks on occluded parts (indicated by orange and yellow circles).
    Our D-ViT captures both semantic image features and the underlying geometric features among landmarks, enabling our model to make more accurate predictions even in the presence of occlusions.
    }
    \label{fig:cofw:pred_blk}
\end{figure}

\begin{figure}[htbp]
\centering
    \captionsetup[subfigure]{labelformat=empty}
    \begin{subfigure}{0.33\linewidth}
        \includegraphics[width=\linewidth]{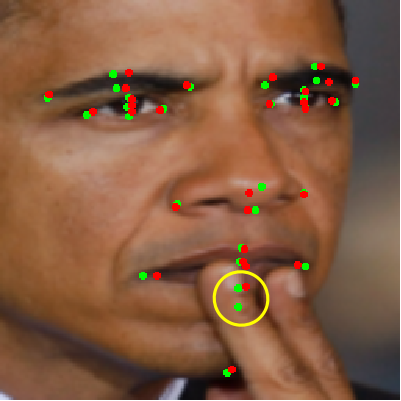}
    \end{subfigure}\hspace{-1mm}
    \begin{subfigure}{0.33\linewidth}
        \includegraphics[width=\linewidth]{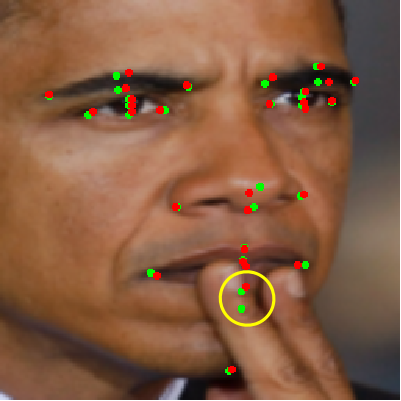}
    \end{subfigure}\hspace{-1mm}
    \begin{subfigure}{0.33\linewidth}
        \includegraphics[width=\linewidth]{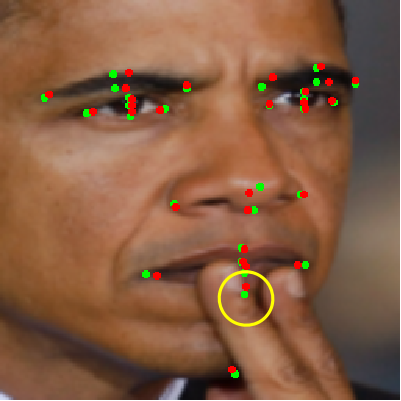}
    \end{subfigure}\vspace{-1mm}
    
    \begin{subfigure}{0.33\linewidth}
        \includegraphics[width=\linewidth]{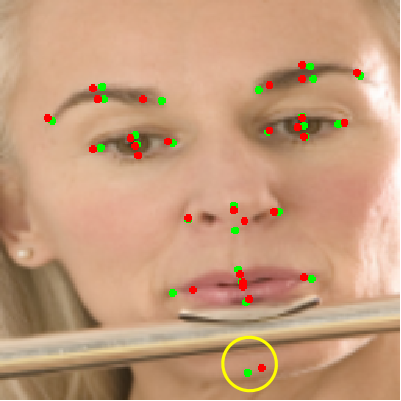}
        \caption{ResCBSP}
    \end{subfigure}\hspace{-1mm}
    \begin{subfigure}{0.33\linewidth}
        \includegraphics[width=\linewidth]{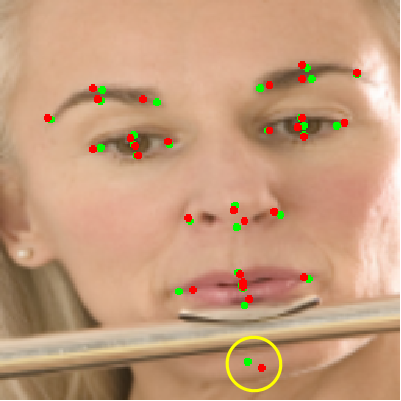}
        \caption{DenC}
    \end{subfigure}\hspace{-1mm}
    \begin{subfigure}{0.33\linewidth}
        \includegraphics[width=\linewidth]{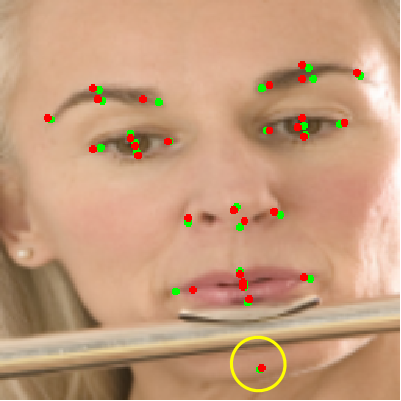}
        \caption{LSC}
    \end{subfigure}
    \caption{Qualitative results of different skip connection strategies on COFW by using 8 prediction blocks. Green and red points represent the predicted and ground-truth landmarks, respectively.}
    \label{fig:cofw:conn}
\end{figure}

\begin{figure}[htbp]
\centering
    \captionsetup[subfigure]{labelformat=empty}

    \begin{subfigure}{0.33\linewidth}
        \includegraphics[width=\linewidth]{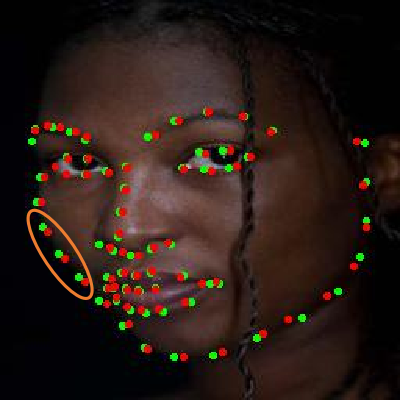}
    \end{subfigure}\hspace{-1mm}
    \begin{subfigure}{0.33\linewidth}
        \includegraphics[width=\linewidth]{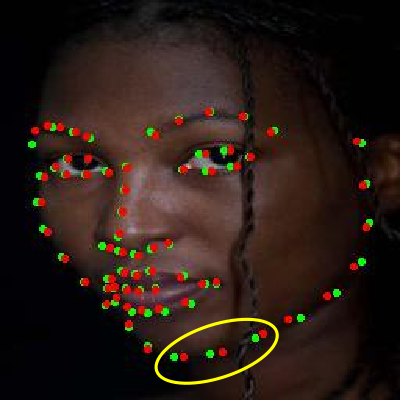}
    \end{subfigure}\hspace{-1mm}
    \begin{subfigure}{0.33\linewidth}
        \includegraphics[width=\linewidth]{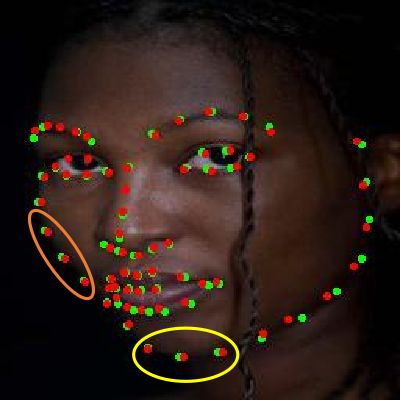}
    \end{subfigure}\vspace{-1mm}
    
    \begin{subfigure}{0.33\linewidth}
        \includegraphics[width=\linewidth]{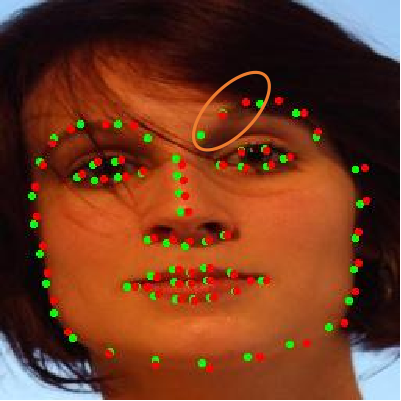}
        \caption{ResCBSP}
    \end{subfigure}\hspace{-1mm}
    \begin{subfigure}{0.33\linewidth}
        \includegraphics[width=\linewidth]{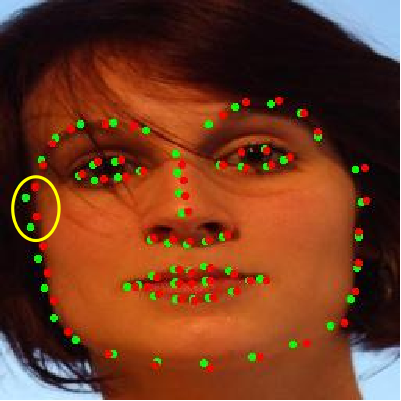}
        \caption{DenC}
    \end{subfigure}\hspace{-1mm}
    \begin{subfigure}{0.33\linewidth}
        \includegraphics[width=\linewidth]{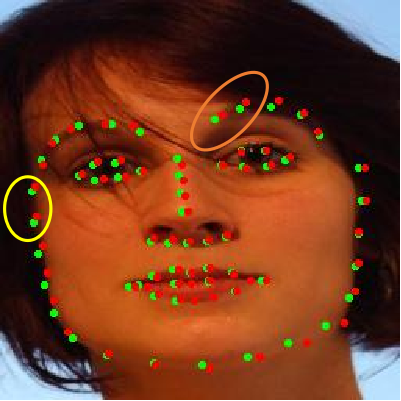}
        \caption{LSC}
    \end{subfigure}
    \caption{Qualitative results of different skip connection strategies on 300W by using 8 prediction blocks. Green and red points represent the predicted and ground-truth landmarks, respectively.}
    \label{fig:300w:conn}
\end{figure}

In this section, we present the qualitative comparison results on COFW and 300W. ~\cref{fig:cofw:pred_blk} and~\cref{fig:300w:pred_blk} show the results of different prediction blocks.~\cref{fig:cofw:conn} and~\cref{fig:300w:conn} show the comparisons of different skip connection strategies.  With the help of our proposed D-ViT and LSC, the detection accuracy for landmarks is improved.

\clearpage

\newpage

\clearpage
{\small
\bibliographystyle{ieee_fullname}
\bibliography{egbib}
}